\documentclass[11pt]{article}

\usepackage[utf8]{inputenc}
\usepackage[T1]{fontenc}
\usepackage[expansion=false]{microtype}
\usepackage[margin=1in]{geometry}
\usepackage{graphicx}
\usepackage{xcolor}
\usepackage{xspace}
\usepackage{booktabs}
\usepackage{amsmath,amssymb,amsthm}
\usepackage{url}
\usepackage[hidelinks]{hyperref}

\usepackage[numbers,sort&compress]{natbib}

\newcommand{\methodname}{\textsc{TauGate}\xspace}
\newcommand{\lora}{\textsc{LoRA}\xspace}
\newcommand{\bitfit}{\textsc{BitFit}\xspace}

\title{Dopamine: Brain Modes, Not Brains\\\large Activation-Gated PEFT via Threshold-and-Gain Tuning}
\author{shervin ghasemlou}

\begin{document}
\maketitle

\begin{abstract}
Parameter-efficient fine-tuning (PEFT) methods such as \lora{} adapt large pretrained models by adding small weight-space updates.
While effective, weight deltas are hard to interpret mechanistically, and they do not directly expose \emph{which} internal computations are reused versus bypassed for a new task.
We explore an alternative view inspired by neuromodulation: adaptation as a change in \emph{mode}---selecting and rescaling existing computations---rather than rewriting the underlying weights.
We propose \methodname{}, a simple activation-space PEFT technique that freezes base weights and learns per-neuron \emph{thresholds} and \emph{gains}.
During training, a smooth gate decides whether a neuron's activation participates; at inference the gate can be hardened to yield explicit conditional computation and neuron-level attributions.

As a proof of concept, we study ``mode specialization'' on MNIST (0$^\circ$) versus rotated MNIST (45$^\circ$).
We pretrain a small MLP on a 50/50 mixture (foundation), freeze its weights, and then specialize to the rotated mode using \methodname{}.
Across seeds, \methodname{} improves rotated accuracy over the frozen baseline while using only a few hundred trainable parameters per layer, and exhibits partial activation sparsity (a minority of units strongly active).
Compared to \lora{}, \methodname{} trades some accuracy for substantially fewer trainable parameters and a more interpretable ``which-neurons-fire'' mechanism.
We discuss limitations, including reduced expressivity when the frozen base lacks features needed for the target mode.
\end{abstract}

\section{Introduction}

Large pretrained models are often adapted to downstream tasks under tight memory and compute budgets.
Parameter-efficient fine-tuning (PEFT) addresses this by updating only a small number of parameters while keeping the base model mostly frozen \citep{Ding_2023}.
Popular approaches include inserting adapters \citep{houlsby2019parameterefficienttransferlearningnlp}, learning continuous prompts \citep{li2021prefixtuningoptimizingcontinuousprompts,lester2021powerscaleparameterefficientprompt}, tuning biases \citep{ben-zaken-etal-2022-bitfit}, scaling internal activations \citep{liu2022fewshotparameterefficientfinetuningbetter}, and adding low-rank weight updates such as \lora{} \citep{hu2021loralowrankadaptationlarge}.

Most PEFT methods are described in \emph{weight space}: they add or update parameters that modify linear maps.
This works well, but it obscures a complementary question that matters for interpretability and efficiency:
\emph{which existing internal computations are reused}, and \emph{which are bypassed}, when a model switches tasks or contexts?
In neuroscience, neuromodulators such as dopamine are often studied as global signals that can change learning and behavior without ``rewiring the brain'' \citep{Schultz_1997}.
A common motif is gain and excitability modulation---shifting when neurons fire and how strongly they contribute---to control network state \citep{Ferguson_2020}.
We use this as an \emph{analogy} to motivate mode switching in networks; we do not claim biological realism.

This paper explores an analogous perspective for PEFT in artificial networks:
adaptation as \textbf{mode switching}---changing \emph{which} neurons participate and \emph{how much} they contribute---rather than learning new weights.
Concretely, we replace low-rank weight deltas with learned activation thresholds (and gains) that gate individual neurons.
The base weights remain frozen; task- or context-specific parameters live entirely in activation space.
We refer to this mechanism as \methodname{}.

\paragraph{Contributions.}
\begin{itemize}
  \item We propose \methodname{}, a lightweight activation-space PEFT method that learns per-neuron thresholds and gains while freezing base weights, enabling explicit conditional computation via neuron-level gating.
  \item We clarify how threshold-based adaptation relates to existing PEFT baselines (e.g., bias tuning and activation scaling) and how ``hard'' gates can yield interpretable, task-specific subnetworks.
  \item We provide a proof-of-concept study on MNIST mode specialization (0$^\circ$ vs 45$^\circ$ rotation) and release a reproducible pipeline that runs on commodity Intel integrated graphics via DirectML.
\end{itemize}

\section{Related Work}

\paragraph{Parameter-efficient fine-tuning (PEFT).}
PEFT techniques reduce the cost of adapting pretrained models by restricting trainable parameters.
Adapters insert small bottleneck modules into each layer \citep{houlsby2019parameterefficienttransferlearningnlp}.
Prompt- and prefix-based methods optimize continuous vectors prepended to the input or intermediate states \citep{li2021prefixtuningoptimizingcontinuousprompts,lester2021powerscaleparameterefficientprompt}.
Other approaches tune small subsets of existing parameters, such as biases (\bitfit{}) \citep{ben-zaken-etal-2022-bitfit}, or rescale internal activations (IA$^3$) \citep{liu2022fewshotparameterefficientfinetuningbetter}.
\lora{} adds low-rank weight updates to linear layers \citep{hu2021loralowrankadaptationlarge} and is widely used in practice.
Surveys summarize the rapidly growing PEFT design space \citep{Ding_2023}.

\paragraph{Activation-space modulation and gating.}
Our work is closest in spirit to activation-space methods such as IA$^3$ \citep{liu2022fewshotparameterefficientfinetuningbetter} and bias tuning \citep{ben-zaken-etal-2022-bitfit}, but differs in \emph{mechanism and intent}:
we explicitly learn per-neuron thresholds that gate activity and can be hardened to yield conditional computation and a ``which-neurons-fire'' interpretation.

\paragraph{Conditional computation and sparsity.}
Mixture-of-Experts (MoE) layers route tokens to expert subnetworks via learned gates \citep{shazeer2017outrageouslylargeneuralnetworks}, and Switch Transformers scale this idea efficiently \citep{fedus2022switchtransformersscalingtrillion}.
\methodname{} can be viewed as conditional computation at neuron granularity: instead of selecting among experts, it selects and rescales individual units within a frozen network.

\paragraph{Neuromodulation as an analogy.}
In biological systems, dopamine is a prominent global signal associated with learning and reward prediction error \citep{Schultz_1997}.
More broadly, neuromodulatory inputs can change neuronal gain and excitability, shifting network state without changing synaptic wiring \citep{Ferguson_2020}.
We borrow this ``mode switching'' intuition to motivate activation-threshold adaptation, while keeping our contributions squarely in the engineering domain.

\section{Method}

\subsection{Problem setup}
Let $f(x; W)$ be a pretrained network with weights $W$.
Given a downstream dataset $\mathcal{D}$, PEFT seeks a small parameter set $\theta$ such that $f(x; W, \theta)$ adapts to $\mathcal{D}$ while keeping $W$ frozen (or mostly frozen) \citep{Ding_2023}.

\subsection{\methodname{}: threshold-and-gain tuning}
The core idea is to move adaptation from \emph{weight space} to \emph{activation space}.
Consider a layer with pre-activation
\begin{equation}
  z_\ell = W_\ell h_{\ell-1} + b_\ell,
\end{equation}
and activation $h_\ell = \phi(z_\ell)$ (e.g., ReLU).
\methodname{} introduces per-neuron thresholds $\tau_\ell$ and gains $\gamma_\ell$ and defines a smooth gate
\begin{equation}
  g_\ell = \sigma\!\left(s \,(z_\ell - \tau_\ell)\right),
\end{equation}
where $\sigma$ is the logistic sigmoid and $s>0$ controls sharpness.
The gated activation is
\begin{equation}
  h_\ell = \left(\gamma_\ell \odot \phi(z_\ell)\right) \odot g_\ell,
  \label{eq:taugate}
\end{equation}
with element-wise product $\odot$.
We freeze $(W_\ell, b_\ell)$ and train only $(\tau_\ell, \gamma_\ell)$ for the target mode/task.

\paragraph{Hard gates at inference.}
At inference time, one can optionally harden the gate to obtain explicit conditional computation:
\begin{equation}
  g_\ell^{\text{hard}} = \mathbf{1}[z_\ell > \tau_\ell],
\end{equation}
yielding a task-specific subnetwork defined by the active units.
This makes ``which units fired'' directly inspectable, and in implementations that can skip zeroed units it can reduce compute.

\paragraph{Regularizing sparsity.}
To encourage sparse conditional computation, we can add a penalty on mean gate activity:
\begin{equation}
  \mathcal{L} = \mathcal{L}_{\text{task}} + \lambda \sum_{\ell} \mathbb{E}_{x \sim \mathcal{D}}\left[\text{mean}(g_\ell(x))\right],
\end{equation}
where $\lambda \ge 0$ trades accuracy for sparsity.

\paragraph{Parameter count and comparison.}
For a network with hidden widths $\{d_\ell\}$, \methodname{} adds $2\sum_\ell d_\ell$ parameters (one threshold and one gain per unit).
This is typically far smaller than updating $W$ directly, and is competitive with common PEFT baselines (e.g., bias-only tuning).
Unlike \lora{}, which modifies how a neuron computes via low-rank weight deltas \citep{hu2021loralowrankadaptationlarge}, \methodname{} primarily changes \emph{which} neurons compute and \emph{how strongly} they contribute.

\section{Theory}

\subsection{Hard-gate limit as subnetwork selection}
Consider the gated activation in Eq.~\eqref{eq:taugate}.
As the gate sharpness $s \to \infty$, the smooth sigmoid gate approaches a step function:
$g_\ell \to \mathbf{1}[z_\ell > \tau_\ell]$ almost everywhere.
In this limit, a fixed input $x$ induces a deterministic \emph{active set} of units at each layer.
The network's forward pass can be viewed as evaluating a task-specific subnetwork obtained by masking inactive units.
This gives a simple mechanistic interpretation: adaptation shifts thresholds so that different subsets of pre-existing features become active.

\subsection{Connection to existing PEFT primitives}
Threshold parameters alone are not a silver bullet: if we define $h_\ell=\phi(z_\ell-\tau_\ell)$ with ReLU $\phi$, then $\tau_\ell$ acts similarly to a learnable bias shift.
This connects \methodname{} to bias tuning (\bitfit{}) \citep{ben-zaken-etal-2022-bitfit}.
Our formulation adds a \emph{multiplicative} gate and a per-neuron gain $\gamma_\ell$, making it closer to explicit conditional computation and to activation-rescaling methods such as IA$^3$ \citep{liu2022fewshotparameterefficientfinetuningbetter}.
The main benefit is not expressivity comparable to full fine-tuning, but an interpretable, neuron-level mechanism that can trade off accuracy for sparsity when the frozen base already contains relevant features.

\section{Experiments}

\subsection{Mode specialization on MNIST}
We study a simple setting designed to highlight \emph{mode specialization} with frozen base weights.
The two modes share labels but differ in input distribution:
(i) standard MNIST (0$^\circ$), and (ii) MNIST rotated by 45$^\circ$.

\paragraph{Foundation pretraining.}
We train a small MLP on a 50/50 mixture of the two modes, producing a single ``foundation'' set of weights.
We then freeze all weights and compare PEFT methods that specialize only by training a small parameter set for the rotated mode.

\paragraph{Model.}
The backbone is a 2-hidden-layer MLP with ReLU activations.
Unless otherwise stated, hidden width is 128.

\paragraph{Baselines.}
We compare against:
(1) \textbf{Frozen} backbone (no adaptation),
(2) \textbf{Full fine-tuning} (all weights trainable),
(3) \bitfit{} (bias-only tuning) \citep{ben-zaken-etal-2022-bitfit},
(4) \lora{} with rank $r{=}8$ \citep{hu2021loralowrankadaptationlarge},
and (5) \methodname{} (threshold-and-gain tuning).

\paragraph{Training and evaluation.}
We report classification accuracy on (a) the 0$^\circ$ test set and (b) the 45$^\circ$ test set, after specializing to the 45$^\circ$ mode.
We run 3 random seeds and report mean $\pm$ standard deviation.
Optimization uses SGD with momentum (to avoid CPU fallbacks in DirectML optimizers), batch size 256, and a short training schedule (2 foundation epochs, 1 adaptation epoch).

\paragraph{Compute.}
All experiments are run on Intel integrated graphics using PyTorch + DirectML (\texttt{torch-directml}) with no discrete GPU available.
Code to reproduce our experiments and regenerate tables/figures is included in \texttt{experiments/run\_mnist\_rotation\_peft.py}.

\section{Results}

\begin{table}[t]
\centering
\caption{Mode specialization on MNIST (0$^\circ$) vs rotated MNIST (45$^\circ$). A small MLP is pretrained on a 50/50 mixture (foundation), then specialized to the rotated mode via PEFT.}
\label{tab:mnist-rotation}
\begin{tabular}{lrrrr}
\toprule
Method & Trainable params & Acc (0$^\circ$) & Acc (45$^\circ$) & High-act frac \\
\midrule
Frozen (no adapt) & 0 & 0.832 $\pm$ 0.017 & 0.833 $\pm$ 0.016 & -- \\
BitFit (bias-only) & 266 & 0.851 $\pm$ 0.005 & 0.852 $\pm$ 0.009 & -- \\
TauGate (threshold tuning) & 512 & 0.850 $\pm$ 0.006 & 0.850 $\pm$ 0.010 & 0.28 \\
LoRA (r=8) & 10448 & 0.855 $\pm$ 0.006 & 0.864 $\pm$ 0.005 & -- \\
Full FT & 118282 & 0.783 $\pm$ 0.007 & 0.887 $\pm$ 0.005 & -- \\
\bottomrule
\end{tabular}
\end{table}

\begin{figure}[t]
\centering
\includegraphics[width=\linewidth]{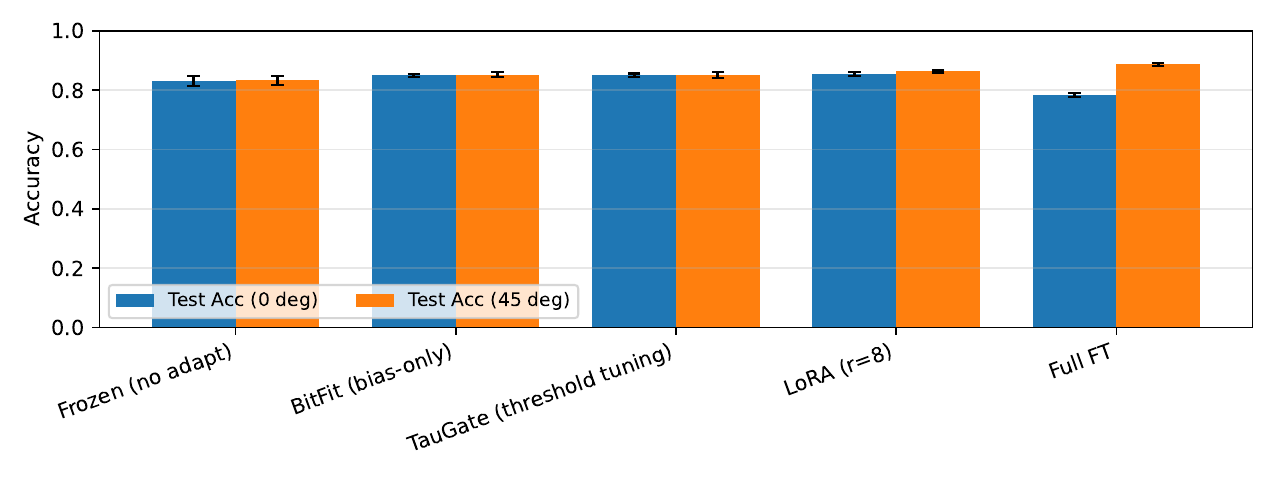}
\caption{Test accuracy on MNIST (0$^\circ$) and rotated MNIST (45$^\circ$) after specializing to the rotated mode.
\methodname{} improves over the frozen foundation baseline with a small parameter budget and exhibits partial activation sparsity (``High-act frac'' in Table~\ref{tab:mnist-rotation}).}
\label{fig:mnist-rotation}
\end{figure}

\paragraph{Accuracy/parameter trade-offs.}
Table~\ref{tab:mnist-rotation} compares PEFT methods under the same frozen foundation backbone.
\methodname{} improves rotated-mode accuracy over the frozen baseline while retaining 0$^\circ$ performance, using only a few hundred trainable parameters (thresholds and gains).
\lora{} attains higher rotated accuracy, at the cost of an order-of-magnitude more trainable parameters.
Full fine-tuning is strongest on the rotated mode but degrades 0$^\circ$ accuracy, reflecting catastrophic interference when weights are rewritten.

\paragraph{Conditional computation signal.}
For \methodname{}, we additionally report the fraction of units with gate value $>0.9$ (averaged across the two hidden layers).
Even in this small setting, specialization concentrates activity into a minority of strongly active units, consistent with the ``mode'' interpretation.
Appendix diagnostics also report gate-mask overlap between 0$^\circ$- and 45$^\circ$-specialized models; in our toy setting the overlap is higher in early layers and lower in deeper layers, suggesting shared low-level features with mode-specific refinement.

\section{Discussion \& Limitations}

\paragraph{When do thresholds help?}
\methodname{} is most natural when the frozen base already contains useful features for the target task/mode, and adaptation is primarily about \emph{selecting} and \emph{reweighting} those features.
In this regime, threshold-and-gain tuning can change the active set of units and yield an interpretable subnetwork view.
However, if the base lacks needed features, gating alone cannot ``invent'' them; weight-space updates (e.g., \lora{} or full fine-tuning) may be necessary.

\paragraph{Relationship to existing PEFT baselines.}
Bias-only tuning (\bitfit{}) \citep{ben-zaken-etal-2022-bitfit} already provides a surprisingly strong and simple PEFT primitive.
Our results suggest that activation-space approaches can be competitive in low-parameter regimes, but also highlight that some benefits depend on the exact setting and backbone capacity.
Unlike pure bias shifts, \methodname{} exposes a gate that can be hardened, giving a direct handle on conditional computation and neuron-level attributions.
The gain component also connects \methodname{} to IA$^3$ \citep{liu2022fewshotparameterefficientfinetuningbetter}.
In the MNIST mode specialization ablation (Appendix), gains account for most of the empirical improvement, while thresholds alone provide little benefit in this toy setting.

\paragraph{Scaling and implementation.}
We present a small-scale proof of concept.
Scaling \methodname{} to large transformers raises practical questions: where to place thresholds (MLP channels, attention outputs, residual streams), how to condition $\tau$ on task/prompt, and how to exploit sparsity for actual wall-clock speedups.
Another challenge is optimization: very sharp gates can create vanishing gradients, while very smooth gates can reduce interpretability.

\paragraph{Dopamine as a metaphor, not a claim.}
We use ``dopamine'' to motivate mode switching and gain/threshold modulation \citep{Schultz_1997,Ferguson_2020}.
We do not claim biological realism; the contribution is a simple engineering mechanism that makes mode-specific computation explicit.

\section{Conclusion}

We proposed \methodname{}, an activation-space alternative to low-rank weight adaptation in which a frozen backbone is specialized by learning per-neuron thresholds and gains.
This framing treats adaptation as \emph{mode switching}: selecting and rescaling existing computations rather than rewriting weights.
In a proof-of-concept MNIST mode specialization setting, \methodname{} improves over a frozen foundation baseline with a small parameter budget and yields an interpretable gating signal, while \lora{} achieves stronger accuracy at higher parameter cost.
Component ablations suggest that, in this toy setting, gain modulation accounts for most of the accuracy improvement, while thresholds primarily provide the gating semantics and conditional-computation handle.

Future work should evaluate threshold-based PEFT on standard transformer benchmarks, study context-conditioned thresholds (prompt- or task-dependent $\tau$), and investigate implementations that exploit hard gates for measurable inference speedups.


\appendix
\section{Additional Details}

\subsection{Reproducibility}
From the project root, create a Python environment and run:
\begin{verbatim}
  .\.venv\Scripts\python.exe .\experiments\run_mnist_rotation_peft.py `
    --device directml --foundation-mix --rotation-degrees 45 --extras
\end{verbatim}
The script writes:
\begin{itemize}
  \item raw results: \texttt{experiments/out/results.csv}
  \item main table: \texttt{tables/mnist\_rotation\_results.tex}
  \item main figure: \texttt{figures/mnist\_rotation\_accuracy.pdf}
  \item extras table (with \texttt{--extras}): \texttt{tables/mnist\_rotation\_ablations.tex}
  \item extras diagnostic (with \texttt{--extras}): \texttt{tables/taugate\_overlap.tex}
\end{itemize}

\subsection{\methodname{} implementation details}
We implement \methodname{} as neuron-wise gating with a sigmoid and a learnable gain:
$h = (\gamma \odot \phi(z)) \odot \sigma(s(z-\tau))$.
The ``High-act frac'' statistic in Table~\ref{tab:mnist-rotation} is computed as the average fraction of gate values exceeding $0.9$ across the two hidden layers on a batch of training examples.

\subsection{Ablations and diagnostics}

\begin{table}[t]
\centering
\caption{Ablations on activation-space adaptation components (mean $\pm$ std over seeds).}
\label{tab:mnist-ablations}
\begin{tabular}{lrrr}
\toprule
Method & Trainable params & Acc (0$^\circ$) & Acc (45$^\circ$) \\
\midrule
Frozen (no adapt) & 0 & 0.832 $\pm$ 0.017 & 0.833 $\pm$ 0.016 \\
GainOnly (activation scales) & 256 & 0.850 $\pm$ 0.005 & 0.849 $\pm$ 0.010 \\
TauOnly (thresholds) & 256 & 0.833 $\pm$ 0.017 & 0.833 $\pm$ 0.015 \\
TauGate (threshold tuning) & 512 & 0.850 $\pm$ 0.006 & 0.850 $\pm$ 0.010 \\
BitFit (bias-only) & 266 & 0.851 $\pm$ 0.005 & 0.852 $\pm$ 0.009 \\
LoRA (r=8) & 10448 & 0.855 $\pm$ 0.006 & 0.864 $\pm$ 0.005 \\
\bottomrule
\end{tabular}
\end{table}

\begin{table}[t]
\centering
\caption{Gate-mask similarity between 0$^\circ$-specialized and 45$^\circ$-specialized \methodname{} models (Jaccard, threshold $g>0.9$; mean $\pm$ std over seeds).}
\label{tab:taugate-overlap}
\begin{tabular}{lrr}
\toprule
Layer & Jaccard & \\
\midrule
Hidden 1 & 1.000 $\pm$ 0.000 \\
Hidden 2 & 0.833 $\pm$ 0.289 \\
Average & 0.917 $\pm$ 0.144 \\
\bottomrule
\end{tabular}
\end{table}

\end{document}